%% file: main.tex
\documentclass[letterpaper, 10 pt, conference]{ieeeconf}  
\usepackage{lipsum}
\usepackage{graphicx}
\usepackage{mwe}   
\usepackage{amssymb}
\usepackage{amsmath}
\usepackage{booktabs}
\usepackage{gensymb}
\usepackage{cite}

\makeatletter
\let\NAT@parse\undefined
\makeatother
\usepackage[bookmarks=true]{hyperref}
\usepackage{mathtools}
\usepackage{etoolbox}
\makeatletter
\patchcmd{\@makecaption}
  {{\footnotesize #1}\\{\footnotesize\scshape #2}}
  {{\footnotesize #1:~\scshape #2}}
  {}{\PackageError{main}{Table caption patch failed}{Check the document class caption definition.}}
\makeatother
\usepackage{xcolor}
\usepackage{listings}
\input{macros.tex}

\IEEEoverridecommandlockouts                              

\title{\LARGE \bf
MIGU: Multimodal Instruction Grounding under Uncertainty \\ for Manipulation Planning
}

\hypersetup{
    colorlinks=true,
    linkcolor=[RGB]{70,130,210},
    citecolor=[RGB]{50, 205, 50},
    urlcolor=[RGB]{70,130,210},
    pdfauthor={None},
    pdftitle={Multimodal Instruction Grounding under Uncertainty for Manipulation Planning},
    pdfsubject={Robotics, Planning},
    pdfkeywords={robotics}
}

\author{ 
Mingke~Lu\textsuperscript{*}, 
Anxing~Xiao\textsuperscript{*}, 
David~Hsu 
\thanks{\textsuperscript{*} denotes equal contribution, alphabetical order 
} 
\thanks{Mingke Lu, Anxing Xiao, and David Hsu are with the School of Computing and Smart Systems Institute, National University of Singapore, Singapore. Correspond to \tt\small anxingxiao@u.nus.edu. 
} 
\thanks{Mingke Lu is also with the Department of Electrical and Computer Engineering, University of California, Los Angeles, Los Angeles, CA, USA.  
} 
} 

\begin{document}
\bstctlcite{IEEEcontrol}

\maketitle
\thispagestyle{empty}
\pagestyle{empty}

\begin{abstract}
Understanding natural human instructions is crucial for deploying robots in human-centric environments. We study multimodal instruction grounding, where language and gesture provide complementary but uncertain cues. We present MIGU, a modular framework that combines semantic and geometric evidence into a unified grounding belief and connects it to manipulation planning. MIGU constructs a 3D geometric likelihood by propagating viewing-direction and depth uncertainty through eye--finger geometry while accounting for hand-direction estimation error. 
A vision--language model (VLM) provides semantic priors over candidate objects and regions, which are combined with the geometric likelihood through Bayes-inspired fusion.
The resulting belief supports behavior planning to either proceed directly to downstream planning or request clarification. Grounded targets then define goals for mobile manipulation and tabletop task-and-motion planning. On a real-world benchmark, MIGU outperforms all evaluated baselines, while ablations support the benefit of explicit multimodal uncertainty modeling. Project website: \href{https://multimodal-instruction.github.io}{multimodal-instruction.github.io}

\end{abstract}

\input{sections/introduction}

\input{sections/relatedworks}
\input{sections/problem}
\input{sections/methods}

\input{sections/exp}

\section{Discussion and Future Work}
\label{sec:discussion}
Our current formulation reflects several practical design compromises. First, it relies on the commonsense reasoning capability of the VLM, assuming that semantic cues do not severely bias the geometric pointing evidence. Second, the semantic confidence provided by the VLM is used as a relative prior rather than a calibrated probability, since reliable calibration is difficult without a huge in-domain calibration dataset. Third, clarification is restricted to a single binary interaction round to meet real-time requirements, rather than being propagated into the manipulation planning process for more tightly coupled decision-making. These choices motivate future work on ensemble-based VLM grounding, calibrated semantic uncertainty, and more tightly integrated uncertainty-aware manipulation planning ~\cite{garrett2020online, curtis2024partially}.

\section{Conclusions}
We presented MIGU, a modular framework that connects multimodal instruction grounding under uncertainty with manipulation planning. MIGU combines a 3D pointing likelihood with a VLM-derived semantic prior to maintain a belief over candidate objects and regions, enabling the robot to choose between direct execution and clarification. On the real-world benchmark, MIGU achieves 83\% grounding accuracy, with ablations confirming the contribution of both modalities. Clarification further improves mobile manipulation planning success from 83.3\% to 91.7\%. Mobile manipulation and tabletop TAMP demonstrations show how grounded references support complex object rearrangement. These results highlight explicit uncertainty modeling as a practical basis for resolving human intent. 

\section*{Acknowledgment}
The authors used OpenAI’s ChatGPT to assist with code writing and manuscript text editing. The authors take full responsibility for the final manuscript and its conclusions.
\bibliographystyle{IEEEtran}
\bibliography{refs}

\end{document}

%% file: macros.tex
\newif\ifcomment
\commenttrue 

\ifcomment

\newcommand{\todo}[1]{\textcolor{red}{TODO: #1}}

\else

\newcommand{\todo}[1]{}

\fi

\definecolor{codegreen}{rgb}{0,0.6,0}
\definecolor{codegray}{rgb}{0.5,0.5,0.5}
\definecolor{codepurple}{rgb}{0.58,0,0.82}
\definecolor{backcolour}{rgb}{0.95,0.95,0.92}

\lstdefinestyle{mystyle}{
    backgroundcolor=\color{backcolour},   
    commentstyle=\color{codegreen},
    keywordstyle=\color{magenta},
    numberstyle=\tiny\color{codegray},
    stringstyle=\color{codepurple},
    basicstyle=\ttfamily\footnotesize,
    breakatwhitespace=false,         
    breaklines=true,                 
    captionpos=b,                    
    keepspaces=true,                 
    numbers=left,                    
    numbersep=5pt,                  
    showspaces=false,                
    showstringspaces=false,
    showtabs=false,                  
    tabsize=2
}

%% file: sections/introduction.tex
\section{INTRODUCTION}

As robots move from structured environments into human-centric spaces such as homes, hospitals, and offices, their role shifts toward assisting and collaborating with people. This requires robots to understand human intent from natural multimodal instructions, including verbal cues and non-verbal gestures such as pointing~\cite{xiao2025robi}. 
However, pointing in the real world is inherently imprecise due to sensing noise and natural variation in human behavior, meaning that the intended object may not lie exactly along the pointing direction~\cite{nickel2007visual,kukier2025empirical}.
Humans can still infer the intended target by combining pointing with language and visual context and seek clarification when uncertainty remains.
How can we equip robots with the same capability?
In this work, we study \textit{uncertainty-aware multimodal instruction grounding} by fusing semantic priors from vision and language with a geometric likelihood derived from pointing geometry and integrating the resulting belief with manipulation planning.

Multimodal instruction grounding remains challenging in open-world deployments. Early learning-based methods jointly learn from labeled multimodal data~\cite{matuszek2014learning,chen2021yourefit,islam2023patron,mane2025ges3vig}, but often generalize poorly out of distribution. With advances in perception models, recent work has adopted more explicit decompositions of language and gesture~\cite{lin2023gesture,gonzalez2026ichores,vanc2024tell,he2026legs}, yet often relies on ad hoc pointing models, struggles with open-ended expressions and continuous-region grounding, and rarely connects seamlessly to manipulation planning.
Recent VLMs show strong visual grounding through direct prediction~\cite{bai2025qwen3,team2025gemini} or visual prompting~\cite{yang2023setofmark,liu2024moka,nasiriany2024pivot}, but when directly applied to multimodal grounding, they still exhibit limited spatial reasoning and can be overconfident.

\begin{figure}[!t]
    \centering\includegraphics[width=0.99\columnwidth]{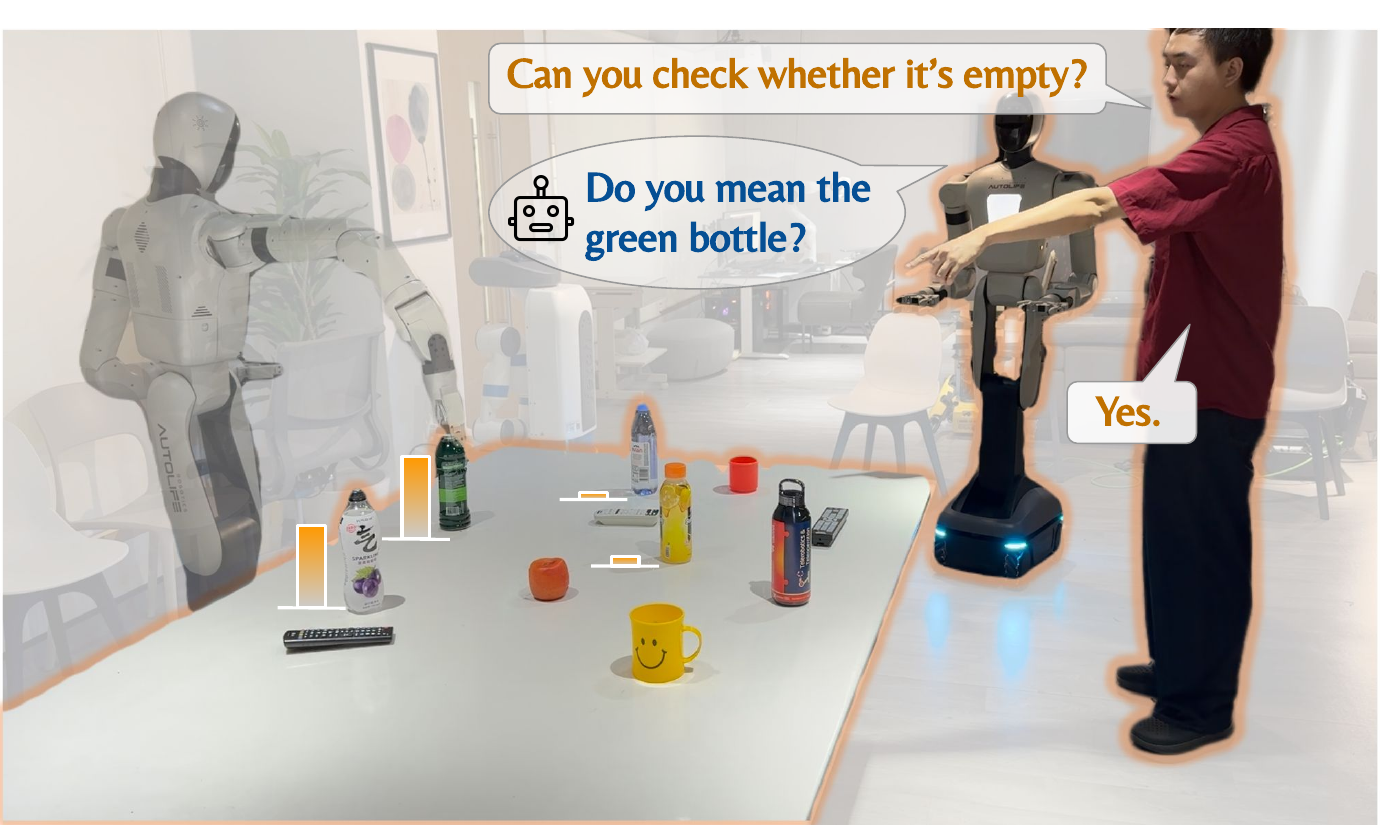}
     \vspace{-0.7cm}
    \caption{MIGU fuses language cues with pointing geometry for robot manipulation under uncertainty, asking clarification questions when needed.}
    \label{fig:cover}
    \vspace{-0.7cm}
\end{figure}

In this work, we explore an explicit and modular approach that exploits the complementary strengths of semantic and geometric cues. We introduce MIGU (\textbf{M}ultimodal \textbf{I}nstruction \textbf{G}rounding under \textbf{U}ncertainty), a framework that combines the geometric information provided by pointing gestures with semantic priors derived from foundation models to obtain a unified belief over objects and regions.
For pointing, we derive a principled 3D geometric likelihood that captures observation, model-fitting, and eye–pointing mismatch uncertainty, yielding a closed-form uncertainty estimate.
For semantic grounding, we study several VLM-based grounding strategies and develop a simple yet effective semantic belief estimator. 
The geometric and semantic beliefs are then fused through Bayes-inspired fusion, producing a unified belief that can be directly used by planning modules. 
We implement an attribute-based behavior planner to balance clarification and direct execution. After clarification, the updated belief is passed to downstream planning, as shown in Fig.~\ref{fig:cover}.

\begin{figure*}[!t]
    \centering
    \setlength{\abovecaptionskip}{0.cm}
    \includegraphics[width=\linewidth]{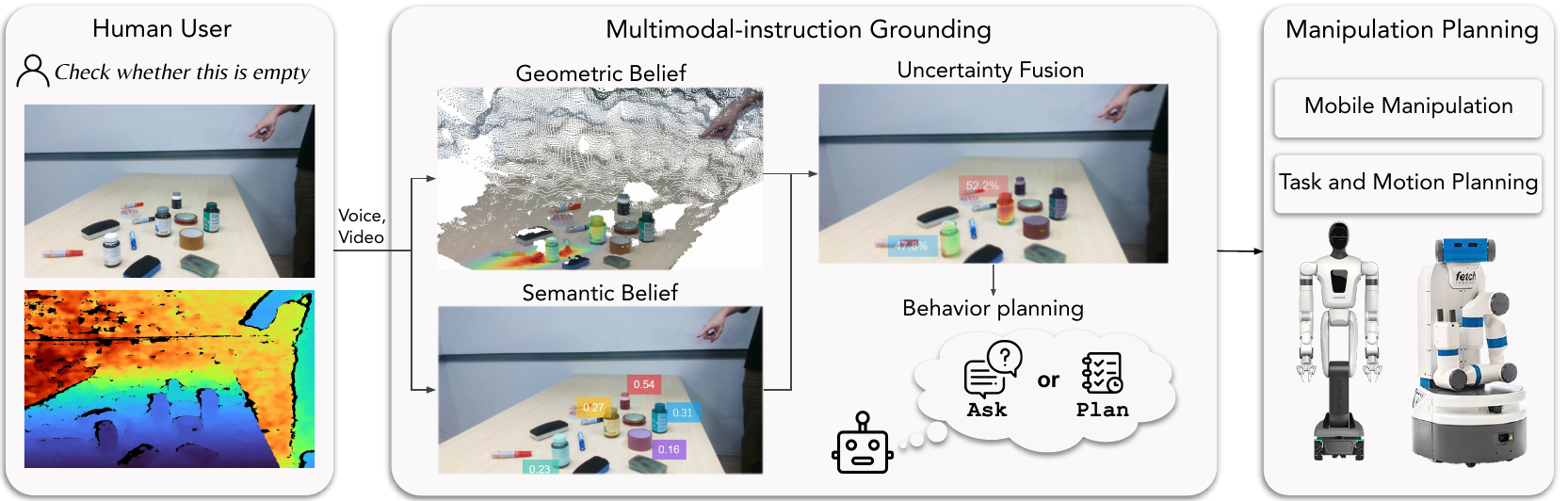}
    \vspace{-0.4cm}
    \caption{\textbf{Framework Overview.} MIGU integrates geometric and semantic evidence into a grounding belief, which is then used for downstream planning.}
    \label{fig:framework}
    \vspace{-0.7cm}
\end{figure*}

We evaluate our approach extensively in real-world multimodal instruction grounding scenarios with diverse environments, objects, referring expressions, and pointing behaviors. Our method outperforms existing baselines, while analysis and ablation studies show that explicitly separating semantic and geometric beliefs provides a more robust basis for multimodal grounding. We further demonstrate how the resulting grounding belief can be integrated into real-world robotic systems, including mobile manipulation and tabletop task-and-motion planning.

%% file: sections/relatedworks.tex
\section{Related Work}

\subsection{Visual Grounding with Language}
Visual grounding connects language instructions to entities in the observed scene~\cite{Cohen2024asurvey}. Probabilistic approaches explicitly model uncertainty over candidate referents~\cite{tellex2011understanding,zhao2023differentiable}, while language-aligned visual representations and open-vocabulary detectors broaden the range of recognizable concepts~\cite{radford2021learning,liu2024grounding,minderer2023scaling}. 
Vision-language models (VLMs)~\cite{bai2025qwen3,team2025gemini} further introduce commonsense reasoning into visual grounding, often through direct prediction or visual prompting techniques~\cite{yang2023setofmark,liu2024moka,nasiriany2024pivot}. However, 3D gesture-grounding data are scarce, and learned models may yield unreliable beliefs at inference time.
For robotic applications, interactive visual grounding has also been explored~\cite{shridhar2020ingress,zhang2021invigorate,yang2022interactive}, enabling robots to actively resolve ambiguity through clarification questions or take additional actions \cite{xiao2026hume} when needed.
Inspired by visual grounding, we adopt a similar belief-based formulation that combines semantic and visual evidence from a VLM with geometric pointing likelihoods in a modular framework. Our method maintains a grounding belief over object and surface references and requests clarification when necessary.

\subsection{Gesture-Informed Grounding}
Pointing helps resolve spatial references, although its interpretation varies with gesture style~\cite{kukier2025empirical} and context~\cite{sauppe2014robot}. Geometric methods estimate pointing directions~\cite{dhingra2020recognition,nakamura2023deepoint} and associate them with candidate objects using rays or cones~\cite{kranstedt2005deixis,nickel2007visual,arslanoglu2025pointing3d}. 
Probabilistic models further capture directional uncertainty from body landmarks in 3D~\cite{pelgrim2024find} or 2D~\cite{he2026legs}. However, such landmarks may be partially or entirely outside the robot camera's field of view.
End-to-end methods~\cite{matuszek2014learning,chen2021yourefit,islam2023patron,mane2025ges3vig} directly infer multimodal grounding, but often generalize poorly out of distribution and offer limited explicit 3D reasoning.
Another line of work focuses on modular fusion of language and pointing, for example by combining object-category evidence with geometric pointing cues for relatively simple instructions~\cite{whitney2016interpreting,vanc2024tell}. LEGS-POMDP~\cite{he2026legs} combines language-target semantic similarity with a probabilistic 2D pointing-cone model. With the rise of foundation models, recent systems use LLMs and VLMs to interpret multimodal instructions and generate executable code~\cite{xiao2025robi,lin2023gesture}. iChores~\cite{gonzalez2026ichores} uses an LLM for instruction interpretation and fuses object detections with a ray-distance-based pointing model~\cite{prochazka2025probabilistic}.
Our approach more fully exploits VLM commonsense and semantic priors while deriving a more principled model of pointing uncertainty, yielding a more accurate grounding belief that benefits behavior planning and downstream manipulation.

%% file: sections/problem.tex
\section{Problem Formulation}

\label{sec:problem}
We consider a robot receiving a multimodal instruction $\mathcal{I}=(L,O)$, where $L$ is a language instruction and $O$ is an RGB-D observation containing a pointing gesture. 
The grounding problem can be formulated as estimating the spatial belief $p(x\mid L, O)$ over candidate target locations $x\in\mathbb{R}^3$. 
We assume that the intended target is represented in the estimated scene. 
For a set of candidate object and region hypotheses $\mathcal{H}$, the corresponding belief over hypotheses is $b(h)=p(h\mid L, O)$, with $h\in\mathcal{H}$.
The robot is equipped with action primitives $\mathcal{A}=\{a_1,\ldots,a_K\}$ and clarification queries $\mathcal{Q}$, each of which elicits a binary response.
Given the grounding belief, it must decide whether to plan and execute the actions or request clarification from the user. The objective is to maximize expected task reward while accounting for clarification costs, subject to the symbolic and geometric constraints required for physical execution.

%% file: sections/methods.tex
\section{Methods}

\subsection{Overview}
MIGU converts a language instruction $L$ and an RGB-D observation $O$ into a grounded task goal for robot execution. As shown in Fig. \ref{fig:framework}, we first model geometric uncertainty to estimate the likelihood of the observed pointing direction for each candidate 3D target. In parallel, a vision-language model identifies candidate objects and regions and assigns semantic confidence scores. Bayes-inspired fusion combines these cues into a belief over hypotheses and locations. A belief-based rule decides whether to execute or request clarification, balancing task reward and interaction cost. 
The selected target then defines the downstream planning goal.

\subsection{Geometric Uncertainty Modeling}

\begin{figure}[!t]
    \centering
    \includegraphics[width=0.93\columnwidth]{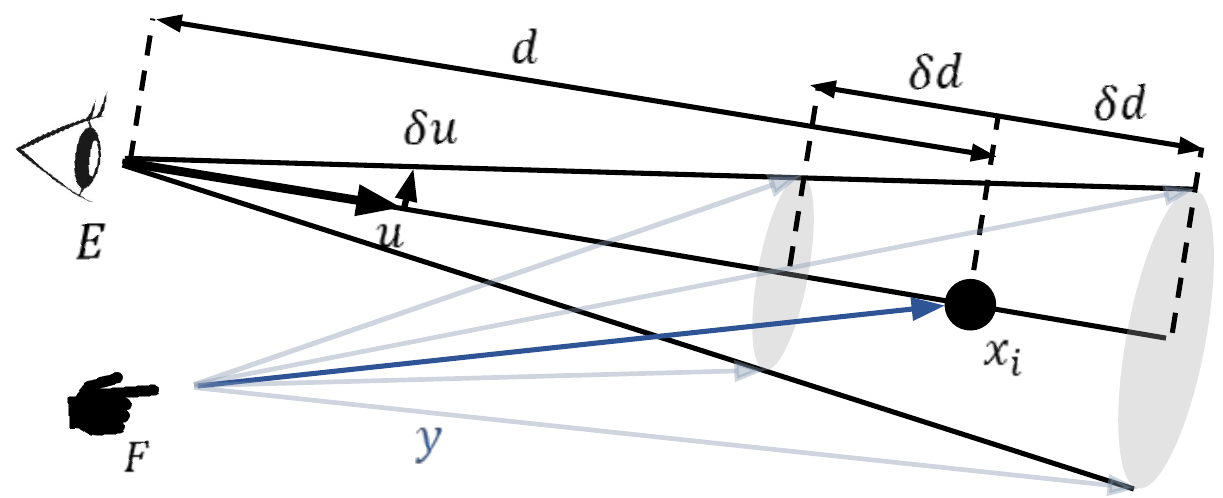}
    \vspace{-0.3cm}
    \caption{\textbf{Gesture Uncertainty Modeling.} Eye--finger geometry propagates viewing-direction and depth uncertainty into the pointing model.}
    \label{fig:gesture_uncertainty}
    \vspace{-0.3cm}
\end{figure}

\begin{figure}[!t]
    \centering    \includegraphics[width=0.93\columnwidth]{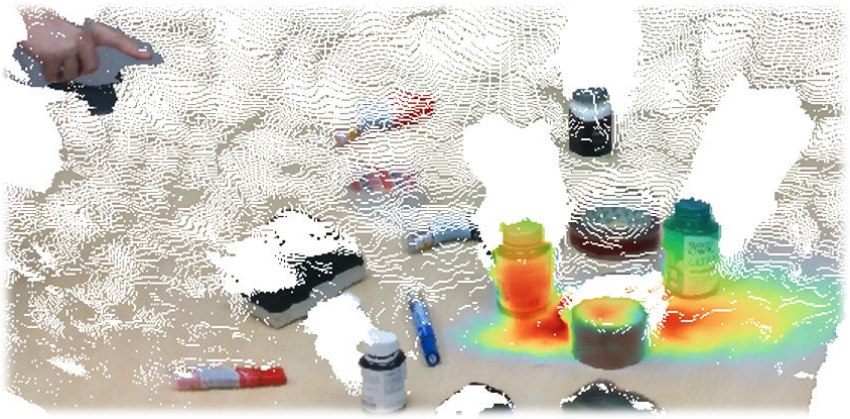}
    \vspace{-0.3cm}
    \caption{\textbf{Illustration of Geometric Likelihood over 3D Scene Points.}}
    \label{fig:gesture_heatmap}
    \vspace{-0.7cm}
\end{figure}

We model gesture uncertainty as arising from the mismatch between viewing and pointing directions, as illustrated in Fig.~\ref{fig:gesture_uncertainty}. 
We assume an average constant eye–hand offset and model how target-perception uncertainty in the viewing direction propagates to the pointing direction.
Let \(E\in\mathbb{R}^3\) denote the eye position, and let
\(x_i\in\mathbb{R}^3\) denote the \(i\)-th 3D point in the scene. We define the nominal viewing distance and unit viewing direction as
\begin{equation}
d_i := \|x_i-E\|_2,
\qquad
u_i := \frac{x_i-E}{d_i}\in\mathbb{S}^2,
\label{eq:gesture_nominal}
\end{equation}
where \(\|\cdot\|_2\) denotes the Euclidean norm in \(\mathbb{R}^3\).
We model uncertainty in the viewing direction locally on the tangent plane
$T_{u_i}\mathbb{S}^2
:=
\left\{v\in\mathbb{R}^3 : u_i^\top v=0\right\}$
and depth uncertainty along the corresponding viewing ray:
\begin{equation}
u\approx u_i+\delta u,
\;
\delta u\sim
\mathcal{N}\!\left(
0,\sigma_u^2P_i
\right),
\;
P_i:=I_3-u_i u_i^\top
\label{eq:visual_direction_uncertainty}
\end{equation}
\begin{equation}
d=d_i+\delta d,
\quad
\delta d\sim
\mathcal{N}\!\left(0,\sigma_{d,i}^2\right),
\quad
\sigma_{d,i}:=\alpha_d d_i,
\label{eq:visual_depth_uncertainty}
\end{equation}
where \(I_3\) is the \(3\times3\) identity matrix,
\(P_i\in\mathbb{R}^{3\times3}\) is the orthogonal projector onto
\(T_{u_i}\mathbb{S}^2\), \(\sigma_u\geq0\) controls the angular uncertainty,
and \(\alpha_d\geq0\) is the relative depth-noise coefficient. Consequently,
\(u_i^\top\delta u=0\) almost surely, and the covariance
\(\sigma_u^2P_i\) represents a rank-two Gaussian distribution supported on
the tangent plane. We assume
$\delta u \perp\!\!\!\perp \delta d.$ 
Based on the visual-localization standard deviations reported by Odegaard et al.~\cite{odegaard2015biases} and the relative standard deviation of visual distance judgments reported by Dukes et al.~\cite{dukes2022visual}, we set
\(\sigma_u=0.044\) rad and \(\alpha_d=0.04\), respectively.

Let \(F\in\mathbb{R}^3\) denote the fingertip position, and 
$y:=E+du-F$
denote the unnormalized finger-to-target vector. A first-order approximation around \((d_i,u_i)\) gives
\begin{equation}
y\mid x_i
\dot{\sim}
\mathcal{N}\!\left(
\bar y_i,\Sigma_{y,i}
\right),
\qquad
\bar y_i=x_i-F,
\label{eq:y_distribution}
\end{equation}
\begin{equation}
\Sigma_{y,i}
=
\sigma_{d,i}^2u_i u_i^\top
+
d_i^2\sigma_u^2
\left(I_3-u_i u_i^\top\right),
\label{eq:y_covariance}
\end{equation}
where \(\dot{\sim}\) denotes equality in distribution under the
first-order approximation.
For \(x_i\neq F\), let
$r_i:=\|\bar y_i\|_2,\;
\mu_i:=\bar y_i/r_i\in\mathbb{S}^2$
denote the nominal finger-to-target distance and direction, respectively.
The Jacobian of the normalization map \(h(y)=y/\|y\|_2\), evaluated at
\(\bar y_i\), is \(J_i\). The corresponding first-order approximation of $h(y)$ around $\bar y_i$ is
\begin{equation}
h(y)\approx\mu_i+J_i(y-\bar y_i),
\quad
J_i
=
\frac{1}{r_i}
\left(I_3-\mu_i\mu_i^\top\right).
\label{eq:normalization_linearization}
\end{equation}

Let \(B_i=[b_{i,1},b_{i,2}]\in\mathbb{R}^{3\times2}\) be an orthonormal basis of the tangent plane \(T_{\mu_i}\mathbb{S}^2\), satisfying
$B_i^\top B_i=I_2,$
and
$B_i^\top\mu_i=0.$
The geometrically induced covariance \(C_i^{\mathrm{geo}}\) and the
overall covariance \(C_i\), both expressed in tangent coordinates, are
\begin{equation}
C_i^{\mathrm{geo}}
=
B_i^\top J_i\Sigma_{y,i}J_i^\top B_i
=
\frac{1}{r_i^2}B_i^\top\Sigma_{y,i}B_i,
\label{eq:tangent_covariance}
\end{equation}
\begin{equation}
C_i
=
C_i^{\mathrm{geo}}
+
B_i^\top
\Sigma^{\mathrm{fit}}
B_i,
\label{eq:geometric_tangent_covariance}
\end{equation}
where \(\Sigma^{\mathrm{fit}}\in\mathbb{R}^{3\times3}\) is the covariance
of the hand-direction estimation error, assumed independent of the
geometric uncertainty, and is projected onto \(T_{\mu_i}\mathbb{S}^2\) and combined with the geometric covariance. 

The discrepancy between the observed direction and the candidate-induced direction is evaluated using the logarithmic map on \(\mathbb{S}^2\).
Let
\(g^{\mathrm{obs}}\in\mathbb{S}^2\) denote the observed unit pointing
direction estimated from the detected hand landmarks.
For
$\theta_i
=
\arccos
\left(\mu_i^\top g^{\mathrm{obs}}
\right),$
the logarithmic map is
\begin{equation}
\operatorname{Log}_{\mu_i}
\left(g^{\mathrm{obs}}\right)
=
\frac{\theta_i}{\sin\theta_i}
\left(
g^{\mathrm{obs}}
-
\cos\theta_i\,\mu_i
\right).
\label{eq:spherical_log_map}
\end{equation}

The corresponding two-dimensional tangent residual is
$z_i
=
B_i^\top
\operatorname{Log}_{\mu_i}
\left(g^{\mathrm{obs}}\right).$
The gesture likelihood is then approximated by a Gaussian distribution in the tangent plane:
\begin{equation}
p\!\left(
g^{\mathrm{obs}}\mid x_i
\right)
\approx
\mathcal{N}\!\left(
z_i;0,C_i
\right)
=
\frac{
\exp\!\left(
-\frac{1}{2}z_i^\top C_i^{-1}z_i
\right)
}{
2\pi |C_i|^{1/2}
}.
\label{eq:gesture_likelihood}
\end{equation}

\subsection{Semantic Uncertainty Modeling and Uncertainty Fusion}
\begin{figure}[!t]
    \centering
    \includegraphics[width=\columnwidth]{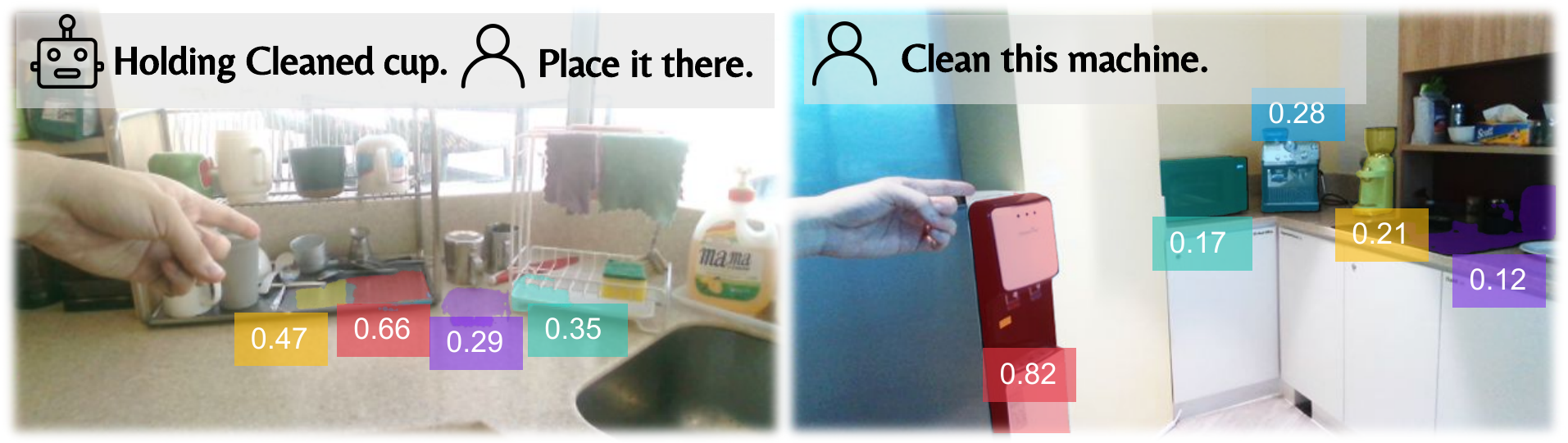}
    \vspace{-0.5cm}
    \caption{\textbf{Generated semantic beliefs.} Left: an ideal prior. Right: a biased prior generated by the VLM.}
    \label{fig:vlm_illustration}
    \vspace{-0.7cm}
\end{figure}
We decompose the multimodal observation \(O\) into the visual scene observation \(O^{\mathrm{vis}}\) and the observed pointing direction \(g^{\mathrm{obs}}\). 
Given \(L\) and \(O^{\mathrm{vis}}\), a VLM
is prompted to identify \(N\) candidate bounding boxes, corresponding
to the hypothesis set
\(\mathcal{H}=\{h_1,\ldots,h_N\}\).
For each hypothesis \(h_i\in\mathcal{H}\), the VLM provides a bounding
box \(\mathcal{B}_i\subset\mathbb{R}^2\) and a confidence score
\(c_i\in[0,1]\).
We make a practical design choice to avoid explicit calibration of such confidence, as it requires a huge calibration dataset.
The VLM jointly reasons over the instruction and scene to infer referents, attributes, affordances, and object roles.
We use GPT-5.6 Sol for its state-of-the-art visual reasoning capabilities.
Each bounding box \(\mathcal{B}_h\) is provided as a spatial prompt to a
class-agnostic segmentation model, which produces the corresponding
pixel-level mask \(M_h\).
Using the RGB-D correspondence, \(M_h\) defines a set of associated 3D scene points
$\mathcal{X}_h
:=
\left\{
x_i\in\mathbb{R}^3
\mid
\pi(x_i)\in M_h
\right\},$
where \(\pi(\cdot)\) denotes projection into the image plane. The VLM confidence induces the following unnormalized semantic score over candidate--point pairs:
\begin{equation}
s_{\mathrm{sem}}(x_i,h\mid L,O^{\mathrm{vis}})
\propto
c_h\,\mathbb{I}\!\left[x_i\in\mathcal{X}_h\right],
\label{eq:semantic_point_prior}
\end{equation}
where \(\mathbb{I}[\cdot]\) is the indicator function. Thus, the semantic confidence of a hypothesis is distributed over all 3D points contained in its mask, without prematurely selecting a particular point.
We assume that the semantic evidence and the observed gesture are conditionally independent given the intended 3D target point, i.e., $p\!\left(g^{\mathrm{obs}} \mid x_i, h, L, O^{\mathrm{vis}}\right)
=
p\!\left(g^{\mathrm{obs}} \mid x_i\right)$.  Motivated by the Bayesian factorization implied by this conditional-
independence assumption, we combine the gesture likelihood and semantic
score into the following unnormalized pointwise fused score:
\begin{equation}
\begin{aligned}
q_h(x_i)
&\propto
p\!\left(g^{\mathrm{obs}}\mid x_i\right)
s_{\mathrm{sem}}(x_i,h\mid L,O^{\mathrm{vis}}) \\
&\propto
p\!\left(g^{\mathrm{obs}}\mid x_i\right)
c_h\,\mathbb{I}\!\left[x_i\in\mathcal{X}_h\right].
\end{aligned}
\label{eq:fused_point_score}
\end{equation}

For each hypothesis, we aggregate its pointwise fused scores using
max pooling, retaining the point that best agrees with the observed
gesture. The resulting hypothesis scores are normalized to obtain an
approximate grounding belief:
\begin{equation}
\tilde b(h)
:=
\max_{x_i\in\mathcal{X}_h} q_h(x_i),
\qquad
b(h)
:=
\frac{\tilde b(h)}
{\sum_{h'\in\mathcal{H}}\tilde b(h')}.
\label{eq:fused_belief_normalization}
\end{equation}

\subsection{Behavior Planning Under Uncertainty}
\label{subsec:planning}
\begin{figure}[!t]
    \centering
    \includegraphics[width=\columnwidth]{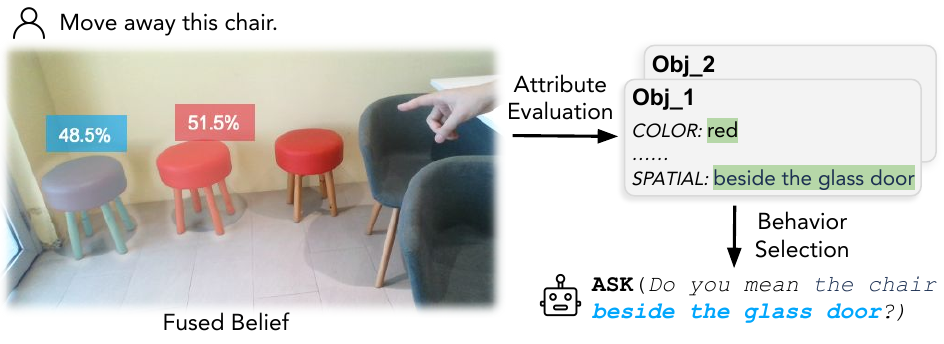}
    \vspace{-0.6cm}
    \caption{\textbf{Behavior Planning.} Attribute-based clarification distinguishes competing targets when the grounding belief remains ambiguous.}
    \label{fig:question_illustration}
    \vspace{-0.7cm}
\end{figure}
We formulate the execute-or-clarify choice as a one-step
belief-space decision problem, equivalent to a one-step POMDP.
To exclude hypotheses with negligible probability, we retain the
smallest set of highest-probability hypotheses whose cumulative belief
is at least 90\%, and renormalize the belief over the retained set.
For simplicity, we continue to denote the resulting
candidate set and belief by \(\mathcal{H}\) and \(b\), respectively.
The robot first considers direct execution of the maximum-a-posteriori hypothesis,
whose expected value is:
\begin{equation}
V_{\mathrm{exec}}(b)
=
b(h^*)R_{\mathrm{success}},
\quad
h^*=\arg\max_{h_i\in\mathcal{H}} b(h_i),
\end{equation}
where \(R_{\mathrm{success}}>0\) denotes the reward for executing the task
with the correct grounding hypothesis.
Alternatively, the robot may request clarification using a binary
attribute-based query. We define a fixed attribute vocabulary
\(\mathcal{C}\) covering visual appearance and spatial relations, such
as color, size, shape, and relative position.
At runtime, the VLM
evaluates these attributes for each candidate hypothesis
\(h_i\in\mathcal{H}\), as visualized in Fig. \ref{fig:question_illustration}. Each attribute that partitions the candidate set
into two nonempty subsets defines a query \(q\in\mathcal{Q}\), together
with the binary predicate
\(\phi_q(h_i)\in\{0,1\}\), which equals one if hypothesis \(h_i\)
satisfies the attribute queried by \(q\), and zero otherwise.
Attributes already specified in the instruction are excluded to avoid redundancy.
Let \(y\in\mathcal{Y}=\{\mathrm{yes},\mathrm{no}\}\) denote the user's
response. Assuming that the response is deterministic and consistent
with the intended hypothesis, its likelihood is
\begin{equation}
p(y\mid h_i,q)
=
\begin{cases}
\phi_q(h_i), & y=\mathrm{yes},\\
1-\phi_q(h_i), & y=\mathrm{no}.
\end{cases}
\label{eq:response_likelihood}
\end{equation}
The response likelihood under the current belief and the corresponding posterior belief are
\begin{equation}
p(y\mid b,q)
=
\sum_{i=1}^{N} p(y\mid h_i,q)b(h_i),
\end{equation}
\begin{equation}
b^{q,y}(h_i)
=
\frac{
p(y\mid h_i,q)b(h_i)
}{
\sum_{j=1}^{N}p(y\mid h_j,q)b(h_j)
}.
\end{equation}

Since only one clarification round is allowed, the robot executes immediately after receiving the response. Let \(C_{\mathrm{query}}(q)\geq 0\) denote the interaction cost of issuing
query \(q\), which may account for communication effort and execution
delay. The value of query $q$ is therefore
\begin{equation}
V_{\mathrm{query}}(b,q)
=
-C_{\mathrm{query}}(q)
+
\sum_{y\in\mathcal{Y}}
p(y\mid b,q)
V_{\mathrm{exec}}\!\left(b^{q,y}\right).
\end{equation}
The optimal query is
$q^*
=
\arg\max_{q\in\mathcal{Q}}
V_{\mathrm{query}}(b,q),$
and clarification is requested iff
$V_{\mathrm{query}}(b,q^*)>V_{\mathrm{exec}}(b).$
The selected query is phrased in the context of the original instruction. 
Otherwise, the robot directly executes $h^*$. This one-step formulation is intentionally lightweight for real-time interaction and excludes multi-round dialogue.

\subsection{System Integration}
\label{subsec:system_integration}

We extract pointing keyframes from synchronized RGB-D images (timestamp offset $\leq80$~ms) using MediaPipe Hands~\cite{zhang2020mediapipe}. Index-finger landmarks 5--8 must form projected segments of at least two pixels, with adjacent direction cosine similarity $\geq0.75$. Fingertip speed must remain $\leq80$~pixels/s for three consecutive observations, with continuity reset after a 0.5~s gap or invalid hand detection. Landmarks are back-projected using calibrated intrinsics and median valid depth in a $5\times5$ pixel neighborhood; observations with invalid depth are discarded. Whisper~\cite{radford2023whisper} transcribes the corresponding language instruction.
After behavior planning, the grounded object and region become fixed arguments of the downstream planning goal. We describe two integrations here.

\begin{figure*}[ht]
\centering
\includegraphics[width=0.95\textwidth]{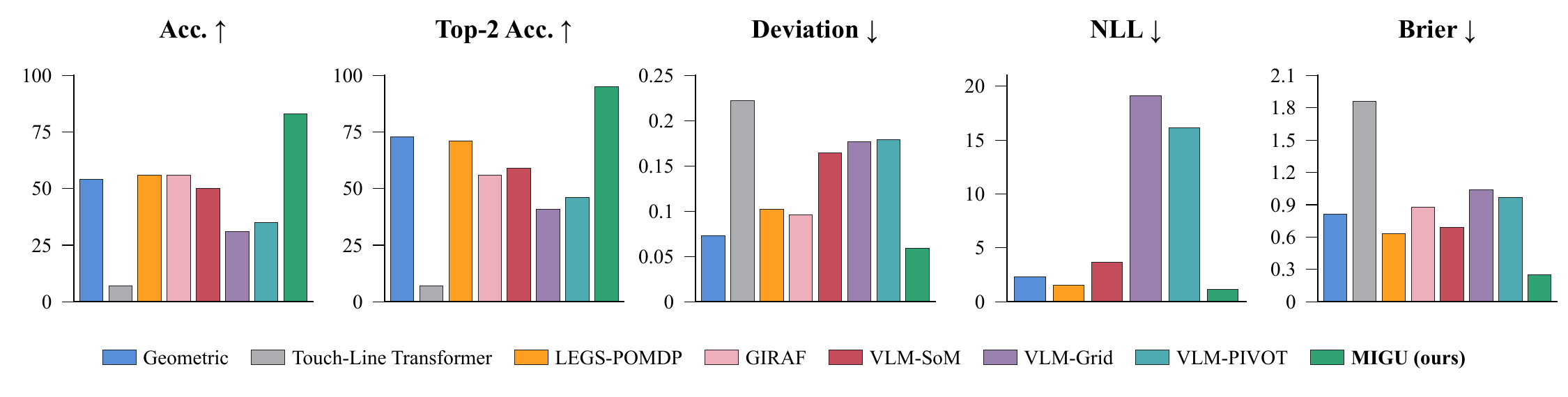}
\vspace{-0.6cm}
\caption{\textbf{Grounding performance against baselines.} MIGU is green; arrows indicate better performance. NLL is omitted for deterministic methods.}
\label{fig:multimodal_grounding_results}
\vspace{-0.5cm}
\end{figure*} 

\noindent\textbf{Mobile Manipulation.}
For mobile picking and placing, the grounded targets specify the object and destination, and we adopt a similar strategy to~\cite{hu2026visibility}. For grasping, we generate 12 heuristic grasp poses and sample candidate base poses from an inverse reachability map~\cite{vahrenkamp2013placement}. For each base--grasp pair, cuRoboV2~\cite{sundaralingam2026curobov2} solves constrained inverse kinematics (IK) in batches. Batched collision checks reject infeasible solutions, and the remaining solutions determine the base-pose ranking. MoveBase navigates to the selected base pose. The robot then observes the target again, generates grasps with Contact-GraspNet~\cite{sundermeyer2021contact}, and solves IK before planning the arm motion with VAMP~\cite{thomason2024motions}. The placement strategy is similar, but uses heuristic sampling of end-effector placements at both near and far distances.

\noindent\textbf{Tabletop Task and Motion Planning.}
We integrate the grounding results with PDDLStream~\cite{garrett2020pddlstream}.
We follow a similar system design as in \cite{curtis2022long}, replacing the object detection with OWLv2~\cite{minderer2023scaling} and the segmentation model with SAM~\cite{kirillov2023segment}.
A grounded object $o$ instantiates a symbolic goal such as \texttt{$on(o_g,o)$}, where $o_g$ is the referred object. A grounded region is represented by a continuous variable $r_g$; for example, ``place the object near here'' translates to \texttt{$near(o, r_g)$}. PDDLStream samples grasps and placements, solves inverse kinematics, and generates collision-free motions to satisfy the goal. The resulting plan is passed to the robot's trajectory-tracking controller for execution.

%% file: sections/exp.tex
\section{Experiments}
\label{sec:experiments}
To better understand the design principles underlying multimodal instruction grounding, we conduct a series of evaluations across diverse real-world daily-life scenarios.

\begin{figure}[h]
    \centering    \includegraphics[width=\columnwidth]{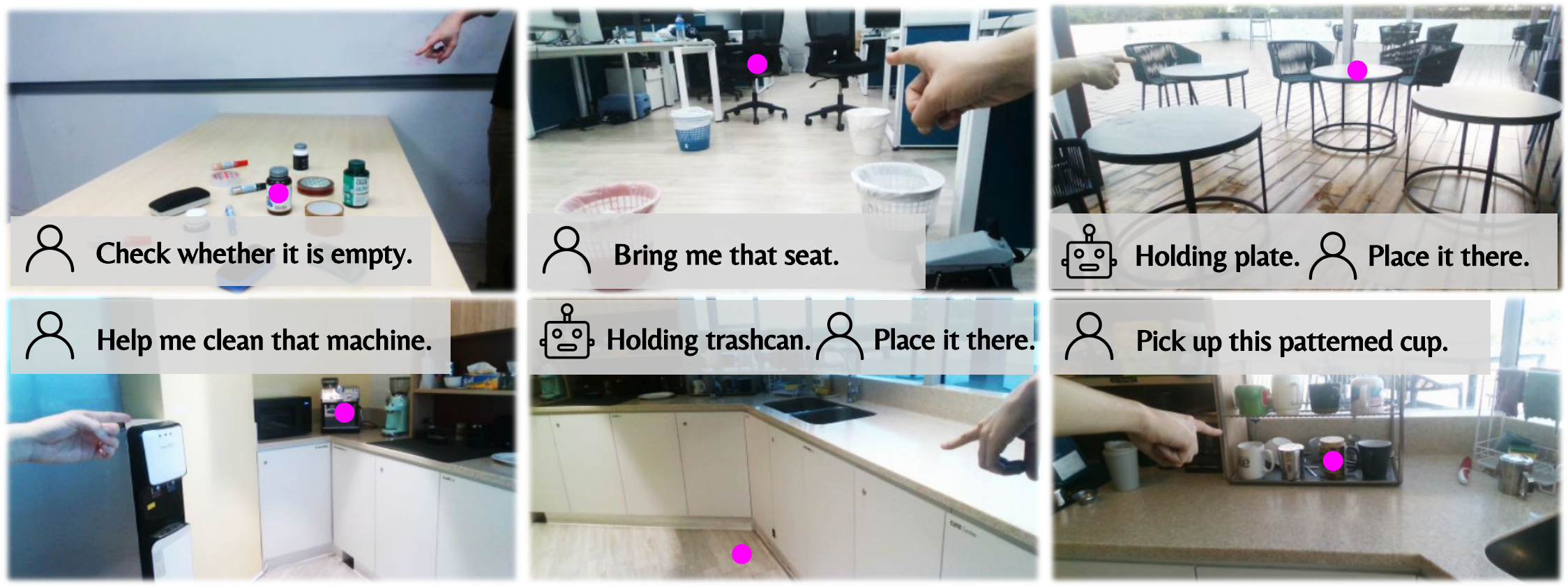}
    \caption{Benchmark examples spanning object and region references. Magenta dots mark annotated targets.}
    \label{fig:benchmarks}
    \vspace{-0.7cm}
\end{figure}

\subsection{Multimodal Instruction Grounding Evaluation}
We first evaluate MIGU’s grounding accuracy and probabilistic prediction quality through comparisons with baseline methods and component-wise ablations.
\subsubsection{Benchmark and Metrics}
We collect 100 multimodal instruction-grounding scenarios, each paired with the language instruction and a ground-truth target, as visualized in Fig.~\ref{fig:benchmarks}. The dataset encompasses semantic ambiguity, geometric ambiguity, and their combination, with interaction distances ranging from short-range tabletop settings (0.2–0.5 m) to medium-range settings (0.5–3 m).
We report \textit{object-level accuracy (Acc.)}, \textit{top-2 accuracy (Top-2)}, \textit{scaled 2D grounding deviation (Deviation.)}, \textit{negative log-likelihood (NLL)}, and \textit{Brier score}. Higher accuracy and lower deviation, \textit{NLL}, and \textit{Brier score} indicate better performance. \textit{NLL} evaluates the probability assigned to the correct referent, while \textit{Brier score} evaluates the complete candidate distribution.

\subsubsection{Baselines}
We compare against a geometric method~\cite{kondaxakis2016temporal}, Touch-Line Transformer~\cite{li2023touchline}, LEGS-POMDP~\cite{he2026legs}, and GIRAF~\cite{lin2023gesture}. Three pure VLM variants use Set-of-Mark prompting (VLM-SoM)~\cite{yang2023setofmark}, image grid visual prompting (VLM-Grid)\cite{liu2024moka}, or iterative visual proposal refinement (VLM-PIVOT)~\cite{nasiriany2024pivot}.
All comparisons concern grounding performance on this benchmark. Touch-Line Transformer and GIRAF are deterministic, so NLL are not applicable.

\subsubsection{Results and Analysis}
As shown in Fig.~\ref{fig:multimodal_grounding_results}, \textit{MIGU outperforms all baseline methods in object-level grounding accuracy}. MIGU also achieves the lowest scaled 2D deviation, demonstrating more accurate fine-grained spatial grounding. This is particularly important for geometrically sensitive tasks such as placing, wiping, and cleaning. \textit{Touch-Line Transformer} performs the worst, likely due to limited generalization from in-domain training, while the pure VLM baseline struggles in geometrically ambiguous cases because of its limited spatial reasoning capability. \textit{LEGS-POMDP} relies primarily on 2D geometric computation and also does not fully exploit the advanced reasoning capabilities of modern VLMs.
In terms of probabilistic prediction, MIGU achieves the lowest reported NLL and Brier score, outperforming LEGS-POMDP by 22.5\% and 60.3\%, respectively. This demonstrates its stronger capability for downstream decision-making under uncertainty.

\begin{table}[t]
\caption{Ablation Study.}
\label{tab:grounding_ablation_results}
\setlength{\tabcolsep}{2pt}
\begin{tabular*}{\columnwidth}{@{\extracolsep{\fill}}lrrrrr@{}}
\toprule
Variant & Acc. $\uparrow$ & Top-2. $\uparrow$ & Deviation $\downarrow$ & NLL $\downarrow$ & Brier $\downarrow$ \\
\midrule
w/o semantics & 51\% & 76\% & 0.0869 & 1.37 & 0.49 \\
w/o geometry & 46\% & 64\% & 0.1746 & 1.76 & 0.57 \\
Geo. $\rightarrow$ ray-distance & 76\% & 89\% & 0.0805 & 2.91 & 0.40 \\
Geo. $\rightarrow$ ray-angle & 58\% & 83\% & 0.1441 & 1.63 & 0.51 \\
Geo. $\rightarrow$ distance-threshold & 74\% & 90\% & 0.0851 & 2.88 & 0.36 \\
Sem. $\rightarrow$ VLM+detector & 60\% & 76\% & 0.0984 & 5.08 & 0.47 \\
Sem. $\rightarrow$ VLM-SoM & 77\% & 85\% & 0.0907 & 3.49 & 0.41 \\
Sem. $\rightarrow$ VLM-Grid & 40\% & 45\% & 0.1287 & 17.66 & 1.02 \\
Sem. $\rightarrow$ VLM-PIVOT & 42\% & 49\% & 0.1237 & 14.67 & 0.93 \\
\midrule
MIGU (full) & \textbf{83\%} & \textbf{95\%} & \textbf{0.0592} & \textbf{1.17} & \textbf{0.25} \\
\bottomrule
\end{tabular*}
\vspace{-0.5cm}
\end{table}

\begin{figure*}[ht]
\centering
\IfFileExists{figures/9_snapshots_mm.pdf}{%
\includegraphics[width=0.96\textwidth]{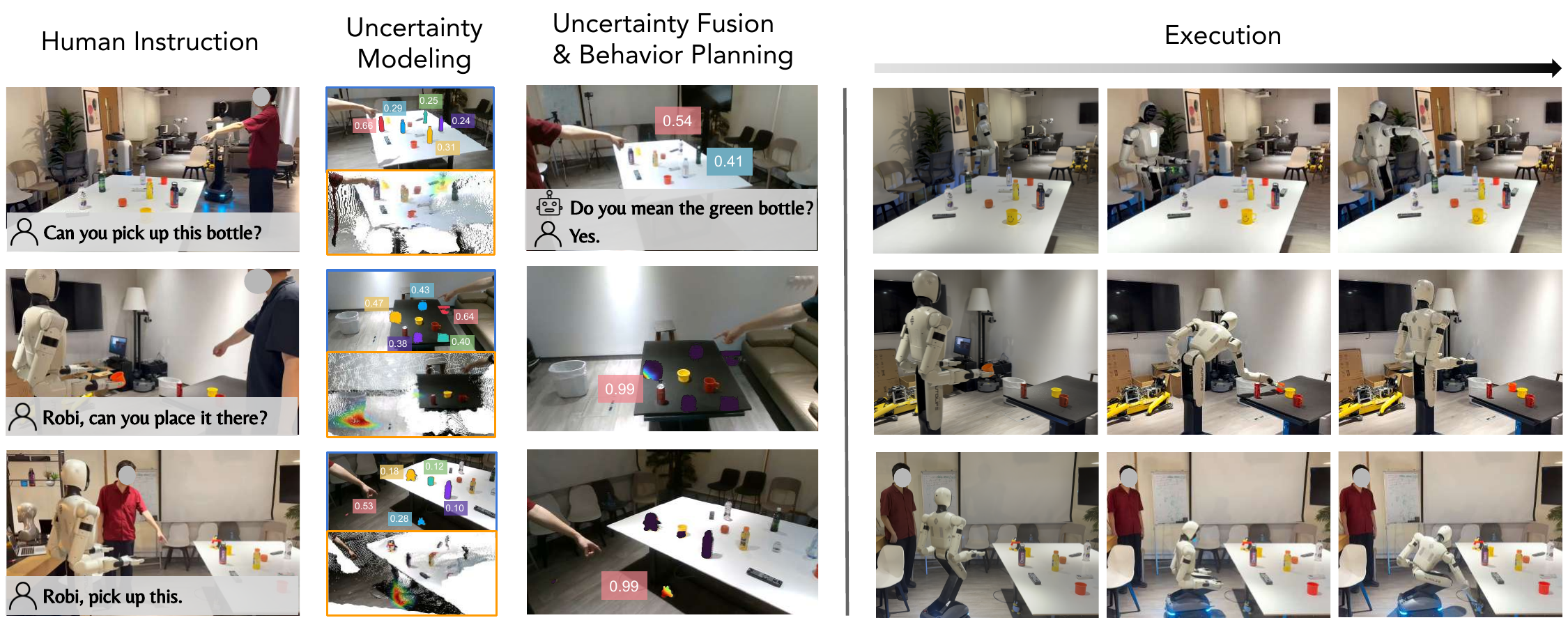}%
}{%
\fbox{\parbox[c][3cm][c]{0.9\textwidth}{\centering Mobile manipulation snapshots pending.}}%
}
\vspace{-0.4cm}
\caption{\textbf{Snapshots of Mobile Manipulation Tasks.} Mobile manipulation from language and pointing, with clarification for ambiguous targets.}
\label{fig:mmshapshots}
\vspace{-0.5cm}
\end{figure*}

\subsubsection{Ablation Study}
Table~\ref{tab:grounding_ablation_results} compares modality removal and component replacement. 
In the variant names, $\rightarrow$ denotes replacing the indicated geometric (Geo.) or semantic (Sem.) component with the method on the right. 
\textit{ray-distance}~\cite{prochazka2025probabilistic} scores distance to the pointing ray, \textit{ray-angle} \cite{whitney2016interpreting} uses a Gaussian angular likelihood with \(1\,\mathrm{rad}\) standard deviation, and \textit{distance-threshold} assigns uniform likelihood within \(0.2\,\mathrm{m}\) of the ray.
 \textit{VLM+detector} uses VLM-generated queries with an open-vocabulary detector to obtain semantic boxes, with a uniform distribution within each box.

Removing semantics or geometry reduces accuracy from 83\% to 51\% and 46\%, respectively, demonstrating the benefit of combining both modalities.
Replacing the proposed geometric model with ray-distance also yields poorer performance in both accuracy and probabilistic metrics, \textit{supporting that the proposed geometric model contributes to better belief modeling and probabilistic prediction}.
We further investigate different strategies for generating the semantic prior. Replacing direct VLM-generated grounding regions with an external detector API, SoM, Grid, or PIVOT consistently degrades performance. A possible explanation is that\textit{ these approaches introduce detector or prompting bottlenecks that lose fine-grained spatial information, whereas directly querying the VLM preserves richer grounding cues}.

\subsection{Integration with Mobile Manipulation}
\label{subsec:exp_mobile_manipulation}
We integrate MIGU with the mobile manipulation system described in Sec.~\ref{subsec:system_integration} on an Autolife S2 wheeled humanoid robot, equipped with an omnidirectional base, a 4-DoF waist, and two 7-DoF arms with grippers. 
We consider eight real-world tasks, each repeated three times. The scenes follow a setup similar to the grounding benchmark, where language instructions and pointing gestures jointly specify manipulation targets. In the clarification-enabled setting, the robot determines whether and how to ask a clarification question as described in Sec.~\ref{subsec:planning}, then updates its grounding belief based on the user’s response. 
We compare the planning success rates (\textit{Planning SR.}) of four variants: semantic-only grounding, geometry-only grounding, and MIGU without and with clarification. 
As shown in Table~\ref{tab:mobile_manipulation_results}, fusing semantic and geometric evidence achieves a success rate of 83.3\%, substantially higher than semantic-only (45.8\%) and geometry-only (41.7\%) grounding. Enabling clarification further improves success to 91.7\%, an increase of 8.4 percentage points. These results demonstrate that both multimodal evidence fusion and clarification-based belief refinement improve the reliability of human multimodal instruction grounding.

Figure~\ref{fig:mmshapshots} illustrates three mobile manipulation tasks.
In the first task (top row), the instruction remains ambiguous among multiple bottles, while sensor noise affects pointing estimation. After fusion, the two leading candidates retain probabilities of 0.54 and 0.41. The robot requests clarification, uses the response to resolve the ambiguity, and plans a whole-body grasp. This case demonstrates the value of clarification when combined semantic and geometric evidence remains inconclusive.
In the second task, the robot holds a cup while the user specifies a placement destination. Geometry alone assigns substantial probability to the floor and locations near the trash can, whereas the semantic prior favors placing the cup on a table. Fusion concentrates the belief on the left-side table region, guiding whole-body placement and illustrating how commonsense reasoning resolves geometric ambiguity.
In the final task, fusion identifies the toy underneath the table with high confidence, allowing execution without clarification. The robot plans a constrained whole-body motion to reach and grasp it, demonstrating the connection between multimodal grounding and manipulation in confined spaces.

\begin{table}[t]
\centering
\caption{Real World Results.}
\label{tab:mobile_manipulation_results}
\footnotesize
\setlength{\tabcolsep}{3pt}
\begin{tabular*}{\columnwidth}{@{\extracolsep{\fill}}lcccc@{}}
\toprule
 & \shortstack{Semantic\\Only} & \shortstack{Geometry\\Only} & \shortstack{Without\\Clarification} & \shortstack{With\\Clarification} \\
\midrule
Planning success & 45.8\% & 41.7\% & 83.3\% & 91.7\% \\
\bottomrule
\end{tabular*}
\end{table}

\begin{figure*}[ht]
\centering
\includegraphics[width=\textwidth]{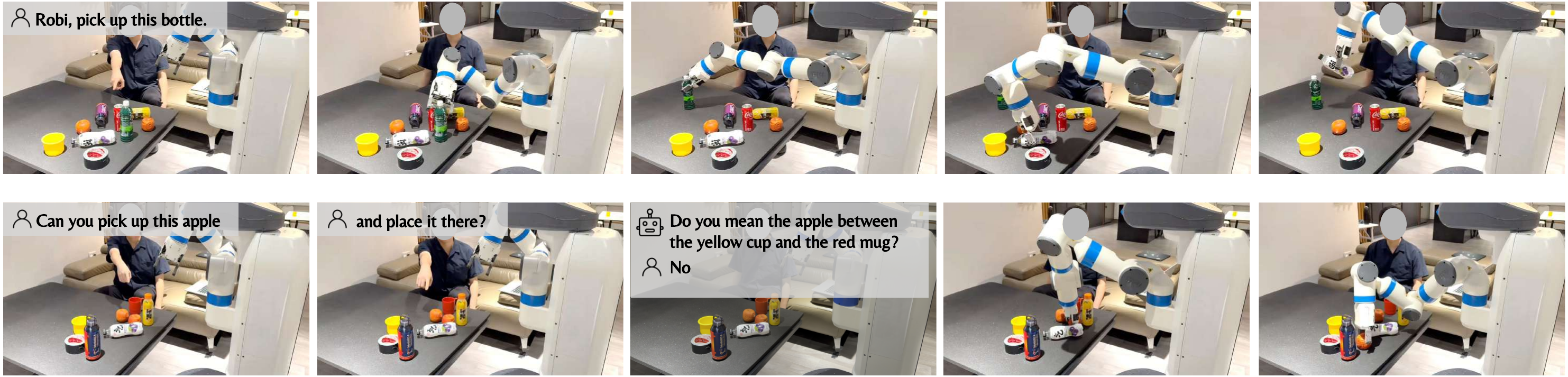}
\vspace{-0.8cm}
\caption{\textbf{Snapshots of Tabletop Manipulation.} Top: obstacle rearrangement and bottle retrieval. Bottom: clarification followed by apple pick-and-place.}
\label{fig:tamp}
\vspace{-0.7cm}
\end{figure*}

\subsection{Integration with Task and Motion Planning}
We demonstrate integration with the TAMP system described in Sec.~\ref{subsec:system_integration} on a Fetch robot~\cite{wise2016fetch} in cluttered tabletop scenes. Language and pointing jointly specify the target object and, for placement tasks, the destination region. The LLM first preprocesses the raw instruction and determines whether the grounding model should be queried twice when a single instruction contains two distinct pointing references.
Figure~\ref{fig:tamp} illustrates two tasks. In the bottle-retrieval task, the robot receives a multimodal instruction and proceeds with planning based on the fused belief without requesting clarification. The solver determines that a blocking bottle must first be removed before retrieving the target, illustrating how a grounded goal can require intermediate manipulation actions.
In the apple pick-and-place task, the robot chooses to ask a spatial clarification question to distinguish between candidate apples. The user's negative response further refines the grounding belief, after which the robot generates a constrained motion plan to complete the task. These examples demonstrate the connection between multimodal reference resolution and task and motion planning, covering both object selection and region specification.